\documentclass[3p]{elsarticle}
\usepackage{hyperref}
\usepackage{subcaption}
\usepackage{gensymb}
\usepackage{todonotes}
\begin{document}
\begin{frontmatter}
\title{PDQ \& TMK + PDQF - A Test Drive of Facebook's Perceptual Hashing Algorithms}
\author{Janis Dalins\corref{cor1}}
\ead{janis.dalins@afp.gov.au;janis.dalins@monash.edu}
\address{Artificial Intelligence for Law Enforcement \& Community Safety (AiLECS) Lab, Australian Federal Police}

\author{Campbell Wilson\corref{cor2}}
\ead{campbell.wilson@monash.edu}
\address{Artificial Intelligence for Law Enforcement \& Community Safety (AiLECS) Lab, Monash University}

\author{Douglas Boudry\corref{cor3}}
\ead{douglas.boudry@afp.gov.au}
\address{Australian Federal Police}

\begin{abstract}
Efficient and reliable automated detection of modified image and multimedia files has long been a challenge for law enforcement, compounded by the harm caused by repeated exposure to psychologically harmful materials. In August 2019 Facebook open-sourced their PDQ and TMK + PDQF algorithms for image and video similarity measurement, respectively. In this report, we review the algorithms' performance on detecting commonly encountered transformations on real-world case data, sourced from contemporary investigations. We also provide a reference implementation to demonstrate the potential application and integration of such algorithms within existing law enforcement systems.
\end{abstract}
\end{frontmatter}

\section{Introduction}
In April 2019 the \textit{Criminal Code Amendment (Sharing of Abhorrent Violent Material) Act 2019} (Cth) criminalised the failure to report and/or remove Australian-related Abhorrent Violent Material (AVM)\footnote{``the most egregious, violent audio, visual material perpetrated by the perpetrator or accomplice''} by ISPs, hosting and content providers\cite{AVM}. Events such as the Christchurch terror attack of March 2019 demonstrated the potential not only for publication of such acts, but also their re-distribution \textit{en masse} after minor modification in order to avoid traditional (cryptographic hash-based) detection and blocking methods by online content providers. The mandatory reporting of such materials by multiple providers worldwide would rapidly overwhelm law enforcement's ability to review and investigate, unless a reliable, portable and acceptable means for measuring similarity is adopted.

In August 2019 Facebook open-sourced the algorithms and released reference implementations of PDQ and TMK + PDQF, their internally used image and video similarity measurement algorithms\cite{FBnews}. At time of writing, these are freely available at the \texttt{ThreatExchange} github repository\footnote{https://github.com/facebook/ThreatExchange/tree/master/hashing}. 

In this report, we run the algorithms through a battery of tests based on typically encountered modification scenarios, using real-world data taken from actual Australian Federal Police (AFP) investigations. Moving beyond mere performance, we also provide a reference implementation of our own, demonstrating the portability and accessibility of such algorithms when effectively packaged and presented through standards-based APIs.

\section{Background}
Cryptographic hashes such as MD5 and SHA-1 have long formed the basis for identifying identical data at the \textit{binary} level. This makes them exceedingly powerful at detecting identical materials - for example, copies of a known file, but by design, completely useless for detecting \textit{similar} materials. Investigations involving large quantities of shared electronic materials (such as AVM and child exploitation materials (CEM)) typically involve light (if not entirely imperceptible) changes to these materials as they are shared. Images and videos, for example, often have watermarks/text added and formats changed (e.g. jpg $\rightarrow$ png), either intentionally by the offenders or as a by-product of the services used to effect sharing and transfer. The efficacy of cryptographic hashes in detecting perceptibly identical materials is therefore limited.

Such limitations would be regarded as inconveniences and inefficiencies in most workplaces, but the true cost of exposure to such materials is only now becoming apparent - not only to front line investigators, but also supporting personnel such as digital forensic examiners and intelligence analysts. Each failure to automatically recognise perceptibly identical materials results in another instance of exposure. Given the sheer volume of materials currently being encountered and seized by Australian police, these repeat exposures could conservatively be estimated in the hundreds of thousands per annum within Australia law enforcement alone. 

Unlike binary-level digest algorithms such as MD5 and SHA-1, perceptual hashes are a mode of fuzzy hashing operating on materials as rendered to the end user, making them highly suitable for detecting lightly or imperceptibly altered materials.

Microsoft has for some years made available PhotoDNA, an image similarity algorithm, on a no-cost basis to law enforcement and other practitioners - primarily for use in detecting CEM. Whilst zero cost, users are required to sign a licensing agreement prior to receiving access to the code, with usage largely limited to organisations dealing directly with CEM. PhotoDNA has been unquestionably effective within child protection, and Microsoft is to be commended for it's early, proactive support in the field. However, the limited rollout and support of this algorithm across law enforcement beyond child protection limits its useability. 

The release and open-sourcing of similarity based hashing algorithms is a welcome development, not only as a means for detecting likely duplicate materials within an organisation, but also for the free sharing of such intelligence at a technical level. Of course, this is on the assumption that they are sufficiently accurate, robust and efficient for deployment and use.

\subsection{The Algorithms}
In this section we provide a very brief overview of the PDQ and TMK + PDQF algorithms. Readers seeking a more in-depth understanding should refer to \texttt{hashing.pdf}\footnote{\url{https://github.com/facebook/ThreatExchange/blob/master/hashing/hashing.pdf}} within the project repository.

\subsubsection{PDQ}
\label{PDQIntro}
In the words of the authors, a \textbf{P}erceptual algorithm utilising a \textbf{D}iscrete Cosine Transform and outputting (amongst others) a \textbf{Q}uality metric, is inspired by and an evolution of the (DCT flavour of) pHash \citep{pHash}, a perceptual hashing algorithm familiar to many within digital forensics.   

PDQ is stored as a 256 bit hash, representable as a 64 character hexadecimal string. We will not attempt to describe the algorithm's design here, but will oversimplify to state that the hash represents the output of a 16x16 transform of the original image via subimage averaging followed by a quantized discrete cosine transform, with each bit being 1 if greater than the overall median in the respective transform position, and 0 otherwise. A quality score broadly proportional to the number of interesting features is introduced to identify image frames that are typically too visually uniform to be useful.

Similarity between two hashes is measured via Hamming distance\footnote{aka 'edit distance' - how many differences there are between two identically lengthed strings/vectors. For example, the Hamming distance between \textbf{abc} and \textbf{abd} is 1. No attempt is made to measure the distance beween changed characters - changing the second vector to \textbf{abz} still results in a Hamming distance of 1.}, with a distance of 0 indicating perceptually/syntactically identical content\footnote{Assuming `quality' imagery}. Statistically, a mean distance of 128 would be anticipated for randomly selected hash pairs, with the authors reporting 30 or less being a good measure for confidently matching materials.

\subsection{TMK + PDQF}
\textbf{TMK + PDQF} (`TMK' for brevity) uses a modified version of PDQ\footnote{The final conversion to binary of each 16x16 transform value is removed, leaving the original figure} for image similarity, combined with the \textbf{T}emporal \textbf{M}atch \textbf{Kernel} for measuring time-related information. Its operation is summarised thus within documentation: (1) resampling videos to a common frame rate (15 frames per second), (2) calculating a frame descriptor, (3) computing averages within various periods and (4) generating a hash from the trigonometrically weighted averages. Given the (in our opinion) greatly increased complexity from PDQ, readers are perhaps best referred to the original paper at \citep{Poullot:2015:TMK:2733373.2806228} for further information.

TMK stores hashes as 258KB binaries (extension \texttt{.tmk}), but the authors are quick to note that the first kilobyte is sufficient to differentiate most videos. This 1KB value (known as the level 1 feature) is an unweighted average of all frame features (i.e. PDQF), and forms the first phase of comparison. The level 2 feature is a collection of weighted averages, totalling 256KB. Metadata regarding the calculation is also included within each file, bringing the total size to 258KB. 

TMK utilises a two phase match comparison process:
\begin{enumerate}
\item{Phase 1: Cosine similarity is calculated between the level 1 (i.e. 1KB) features of two hashes, with a score between -1.0 and 1.0 (inclusive) generated. Higher scores indicate greater similarity. A match threshold of 0.7 is recommended by the authors, and is used throughout our testing.}
\item{Phase 2: The level 2 (i.e. 256KB) features of two hashes are compared and a score in the range 0.0-1.0 generated. Higher scores indicate closer matches, with 0.7 being recommended as a match threshold. This recommendation is adopted within our testing}
\end{enumerate}

\section{The Test Environment}
All tests and activities conducted within this report were carried out on a Dell Precision 5530 notebook computer running Ubuntu Desktop 18.04, installed in August 2019 and patched (using \texttt{apt get update}) on the day of commencing these experiments. 
The PDQ and TMK algorithms and associated software were obtained via a \texttt{git clone} of the Facebook ThreatExchange repository \cite{Fa19}. We then installed \texttt{ffmpeg}, a pre-requisite for TMK and performed \texttt{make} commands for the \texttt{C++} versions \footnote{A python library has since been included within the repository} of both algorithms. The make process includes sense-checks and tests at conclusion - the TMK process reported errors, but these were caused by an incorrect path setting for the ffmpeg binary. We manually performed the tests and observed nil reported errors.

Beyond this minor hiccup, the installation process was fast and uncomplicated, though we recommend users have at least a basic understanding of software compilation and builds for this stage.

\subsection{Test Corpus}
All tests reported within this paper were conducted on a corpus of 225,887 images and 3,366 videos manually reviewed by Police and annotated as child exploitation materials. These files were sourced exclusively from 10 investigations undergoing finalisation by one Australian jurisdiction's Joint Anti Child Exploitation Team (JACET) in July 2019 - the relative skew between media types reflects the materials as located. These matters may remain subject to further judicial proceedings, therefore no further details of their provenance will be provided.

\begin{figure}[h]
\includegraphics[width=\columnwidth]{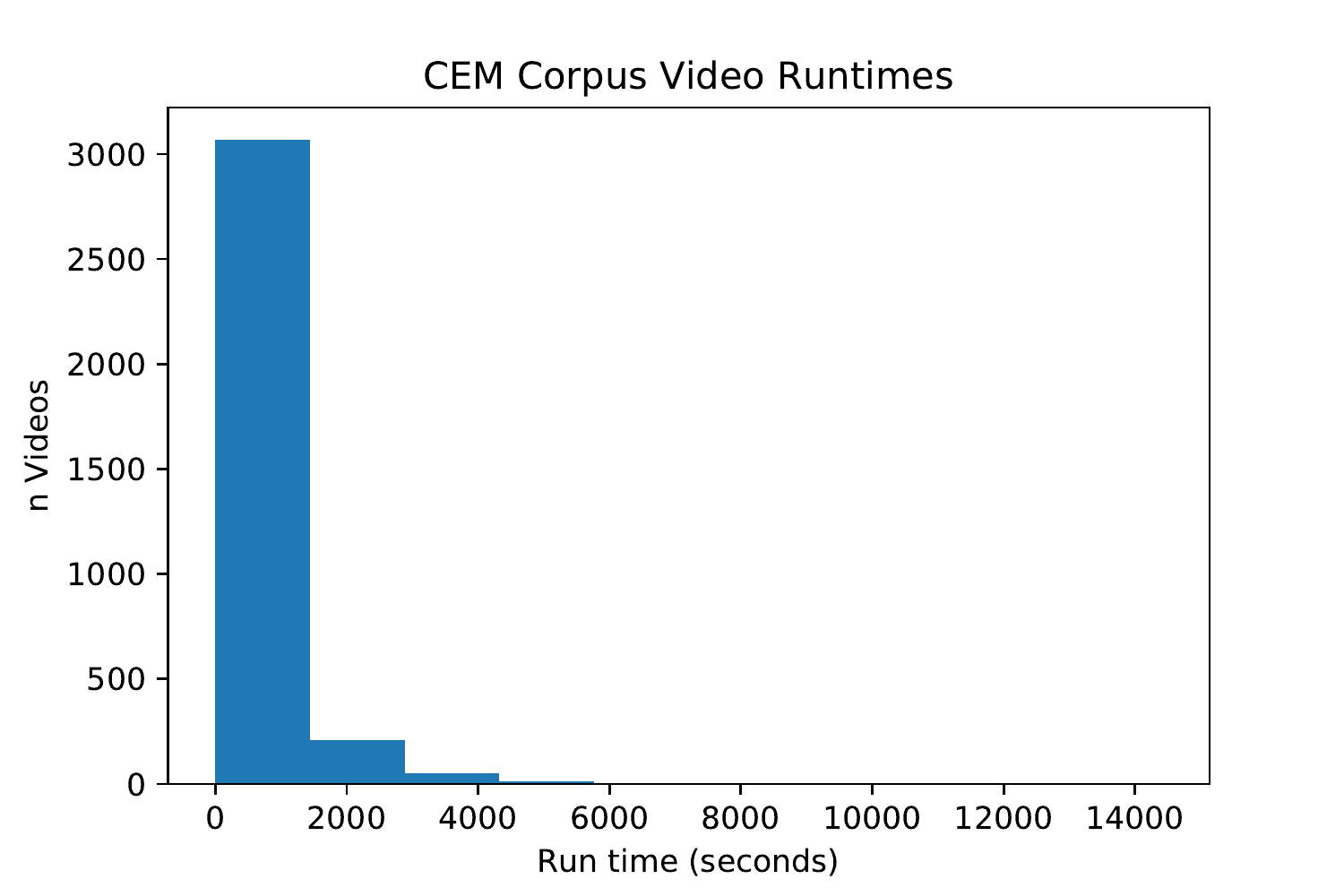}
\caption{Test corpus video runtimes}
\label{CEMRuntimes}
\end{figure}

For tests benefitting from separate corpora, we utilised about 322,490 images of lawful pornography first used by \cite{DALINS201840}, plus a partial (approx. 596,000 images) download of the Google Open Images Dataset \cite{OpenImages}.

\section{Testing Imagery (PDQ)}
Once installed and built, the PDQ software is presented as a series of executables - we utilised \texttt{pdq-photo-hasher} for our tests.
 
We developed a python script automating test corpus (refer Section \ref{pdqmeth}) generation, copying each file from a source drive, calling the pdq hashing binary for each variation, parseing the results (via regex) and then calculating the respective hamming distances.

\subsection{Methodology}
\label{pdqmeth}
\begin{figure*}
  \centering
  \begin{subfigure}{.4\columnwidth}
    \includegraphics[width=\columnwidth]{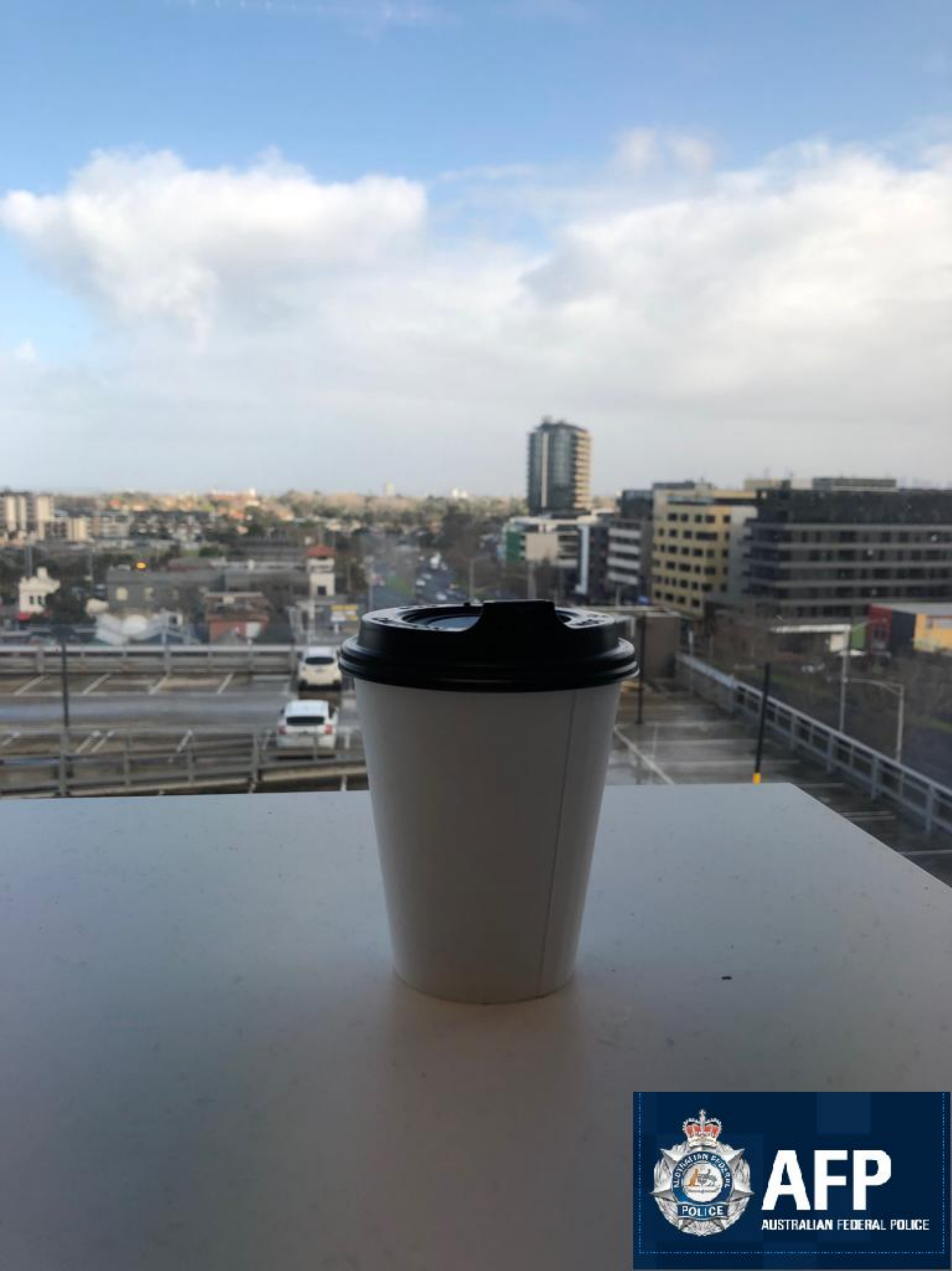}
    \caption{Watermark}
    \label{sampleWatermark}
  \end{subfigure}
  \begin{subfigure}{.4\columnwidth}
    \includegraphics[width=\columnwidth]{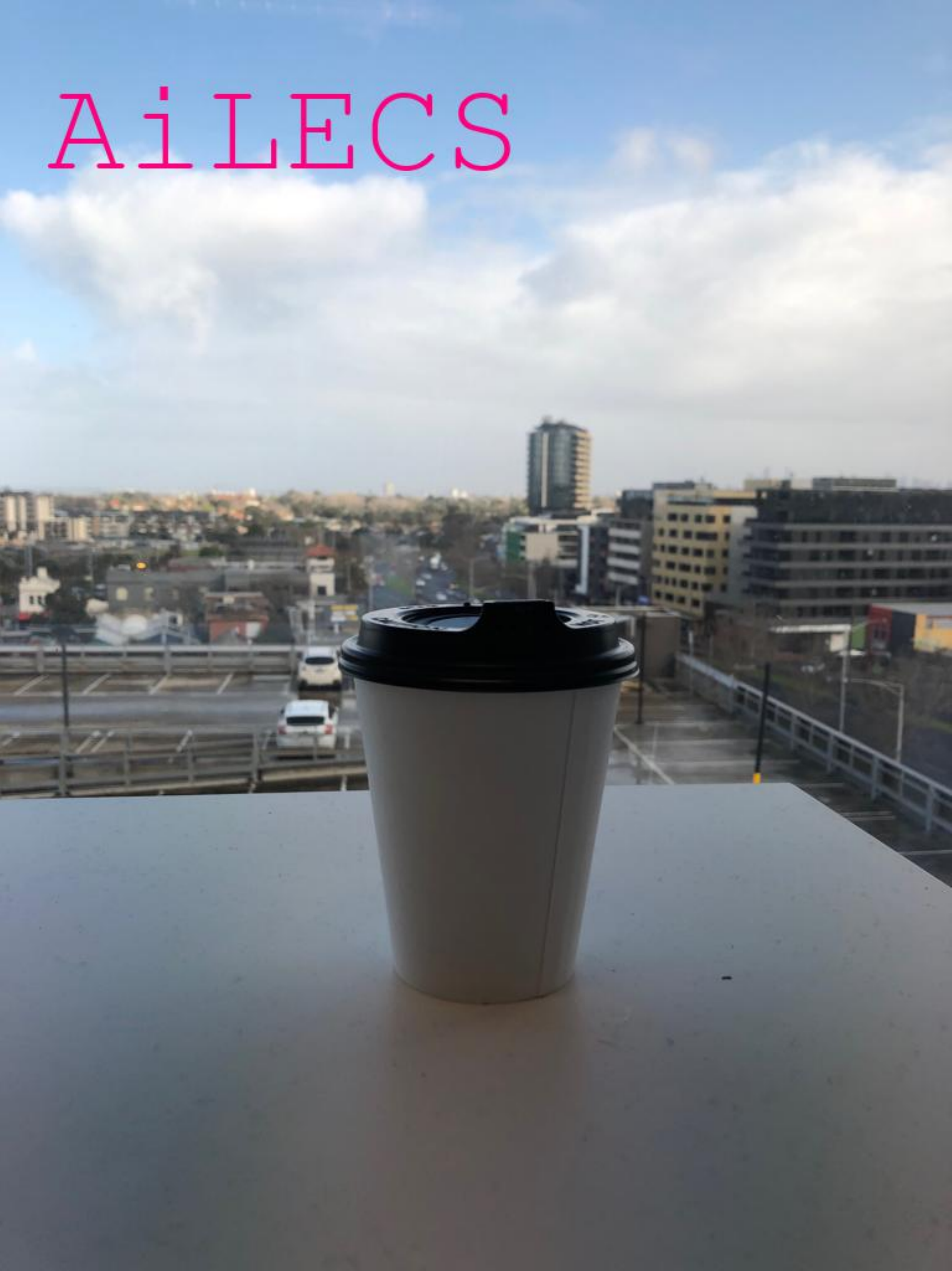}
    \caption{Text}
        \label{sampleText}
  \end{subfigure}
    \begin{subfigure}{.4\columnwidth}
    \includegraphics[width=\columnwidth]{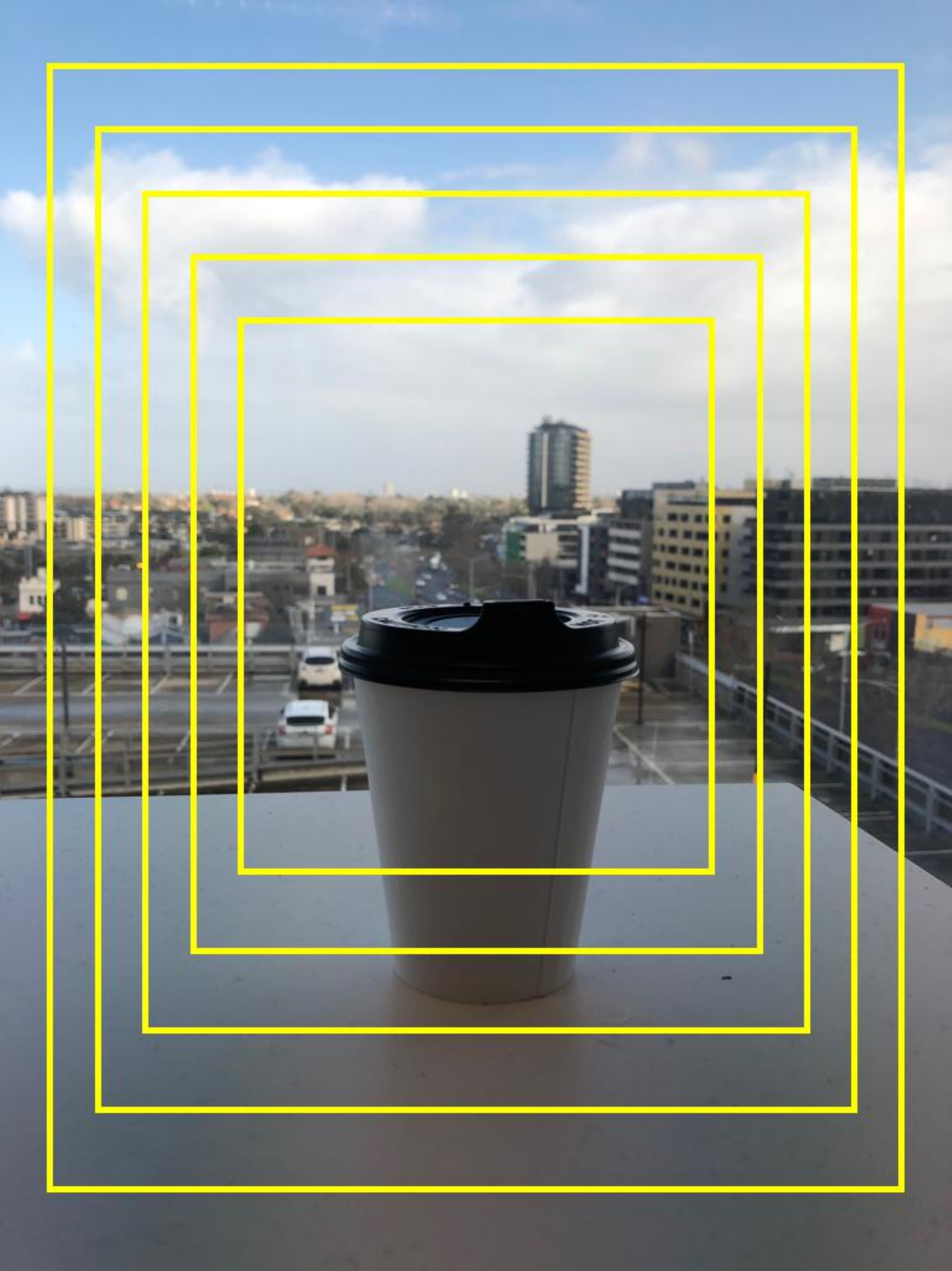}
    \caption{Cropping}
        \label{sampleCropping}
  \end{subfigure}
    \begin{subfigure}{.4\columnwidth}
    \includegraphics[width=\columnwidth]{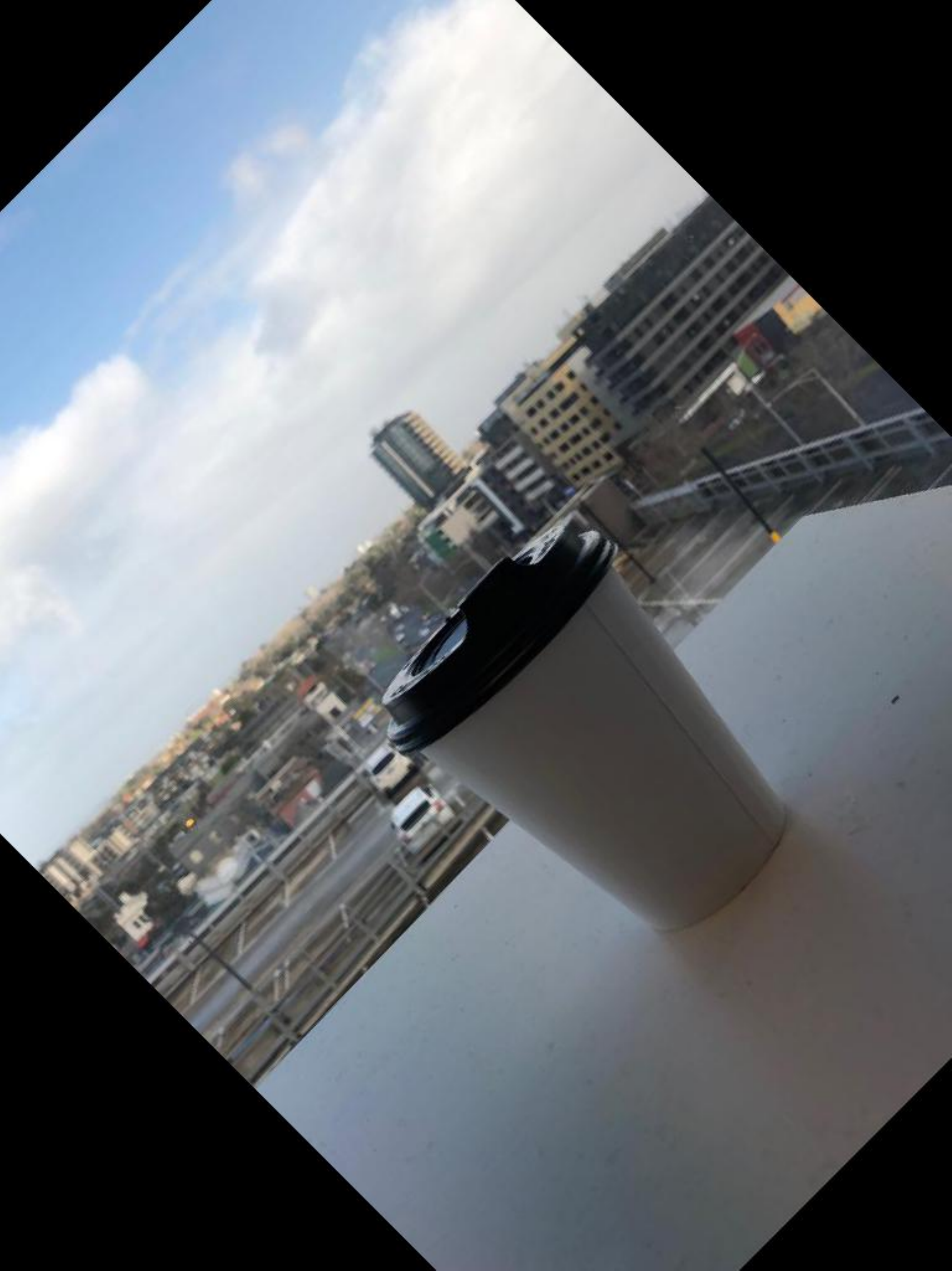}
    \caption{Rotation}
        \label{sampleRotation}
  \end{subfigure}
      \begin{subfigure}{.4\columnwidth}
    \includegraphics[width=\columnwidth]{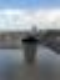}
    \caption{Thumbnail}
        \label{sampleThumbnail}
  \end{subfigure}
  \caption{Image treatments for test corpus. Format change not shown.}
  \label{fig:treatments}
\end{figure*}

Each image within our test corpus was subjected to the following transformations, as illustrated in Figure \ref{fig:treatments}
\begin{enumerate}
\item{Format change - the image is converted to a random selection of JPEG, TIFF, PNG, or bitmap formats.}
\item{Watermark - The AFP logo was added to the bottom right corner, being $\frac{1}{4}$ the size of the image's shorter dimension. The logo is non-transparent, with numerous colour shifts and rendered surfaces.}
\item{Text - `AiLECS' is added to the image in the top left-hand corner, with the font size being the largest before the text itself extends beyond $\frac{1}{2}$ the size of the shorter dimension.}
\item{Thumbnail - the image is reduced in size to 32, 64, 128 and 256 pixels for the longer dimension, akin to file previews automatically generated in applications such as Windows Explorer and MacOS Finder. Figure \ref{sampleThumbnail} displays a 32 pixel thumbnail.}
\item{Cropping - a random proportion of the image greater than 0\% and less than 100\% is selected and retained, using the existing image centre and aspect ratio. Figure \ref{sampleCropping} displays cropping areas for ratios 0.5-0.9 (i.e. retain half $\rightarrow$ retain 90\%) inclusive, using increments of 0.1}
\item{Rotation - the image is randomly rotated between 1 and 359 degrees inclusive. Figure \ref{sampleRotation} displays a 45\degree rotation. }
\end{enumerate}

In order to compare processing overheads, we timed the original hashing of each image using the PDQ algorithm, and also MD5. In order to maintain consistency with the PDQ process, the MD5 was calculated using an external binary (\texttt{md5sum}) rather than from a buffer within the python script. Furthermore, this step was carried out at random either at the first encounter of each file, or later in the process in order to compensate for any operating system or device-level caching. 

\subsection{Results}
We observed the PDQ algorithm to perform strongly when minor/imperceptible changes are made or features are \textit{added} to the rendered image, whilst struggling with removal or alteration.

\subsection{Format Change}
\begin{figure}[h]
\includegraphics[width=\columnwidth]{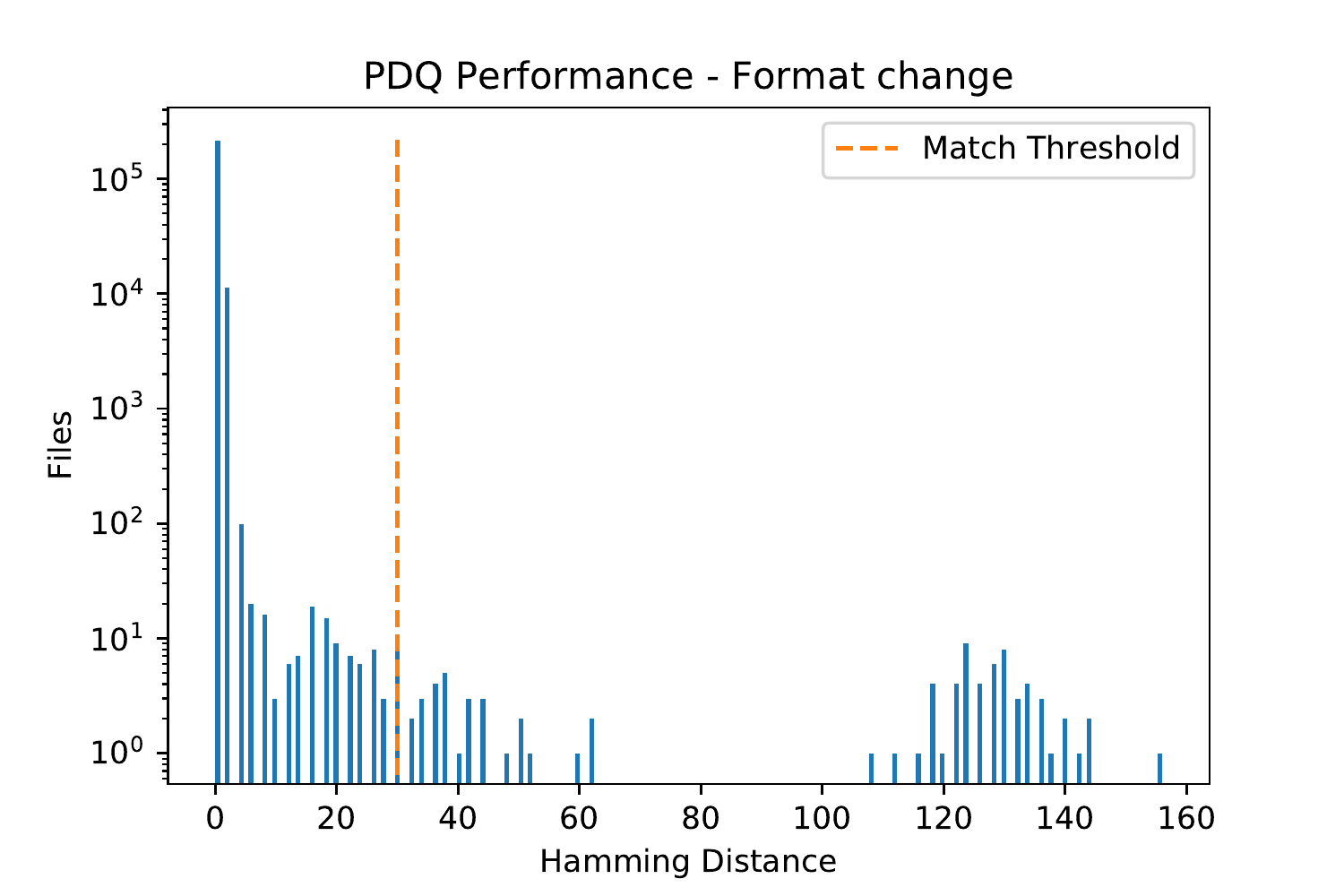}
\caption{Hamming distances resulting from format changes - note use of log scale}
\label{PDQFormatChanges}
\end{figure}

As anticipated, the algorithm was robust to format changes, reflecting its use of the rendered image (rather than underlying binary values) for hash calculation. 99.96\% of format changed images met the match threshold of 30, necessitating the use of log scale for the y axis in Figure \ref{PDQFormatChanges}. To summarise, over 94.85\% of format changes resulted in no hash alteration (i.e. a hamming distance of 0), with a further 5.01\% reporting a hamming distance of 2. No hamming distances of 1 were identified, in keeping with the reasoning detailed in Section \ref{PDQIntro}). 

\subsection{Watermark}
\begin{figure}[h]
\includegraphics[width=\columnwidth]{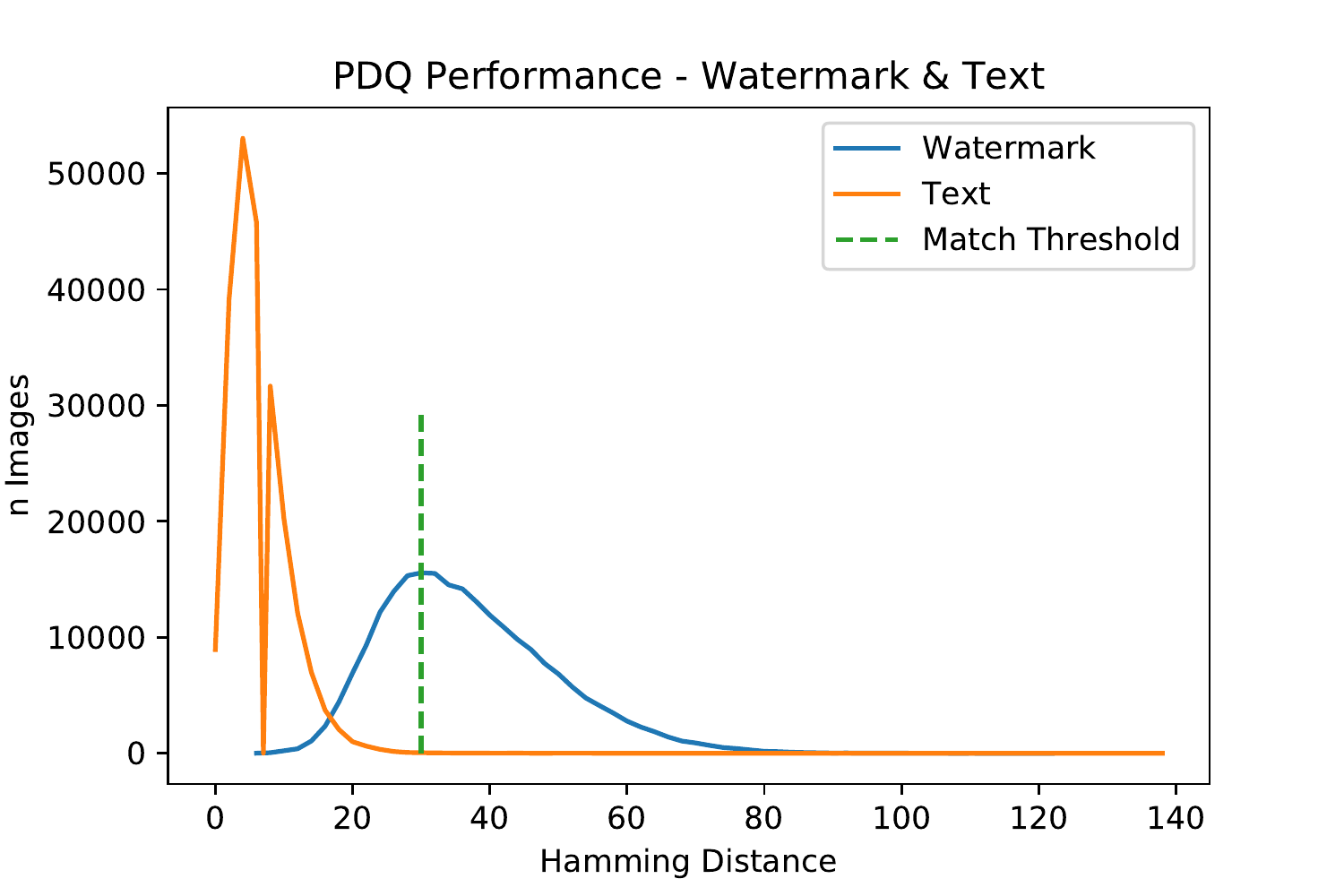}
\caption{Hamming distances resulting text and watermark introductions}
\label{PDQWatermarkText}
\end{figure}

PDQ was not as robust to the introduction of a watermark as it was to changes to underlying formats. Figure \ref{PDQWatermarkText} shows that just overhalf of all watermarked images would be missed if the hamming distance threshold of 30 was strictly adhered to. Whilst disappointing, the test does involve a rather intrusive change to the image, with a completely opaque watermark featuring numerous features and color shifts. This should be regarded as an extreme example of such alteration. 

\subsection{Text}
Being less intrusive than the aforementioned watermarking, PDQ performed more strongly in this test. Figure \ref{PDQWatermarkText} shows that a vast majority of text changes resulted in hamming distances less than 20, with most less than 10. On a side note, the strong dips around 1 and 7 on the x axis most likely reflect the algorithm's bias to returning even hamming distances (as documented within the project) rather than any statistical error.

\subsection{Thumbnail} 
\begin{figure}[h]
\includegraphics[width=\columnwidth]{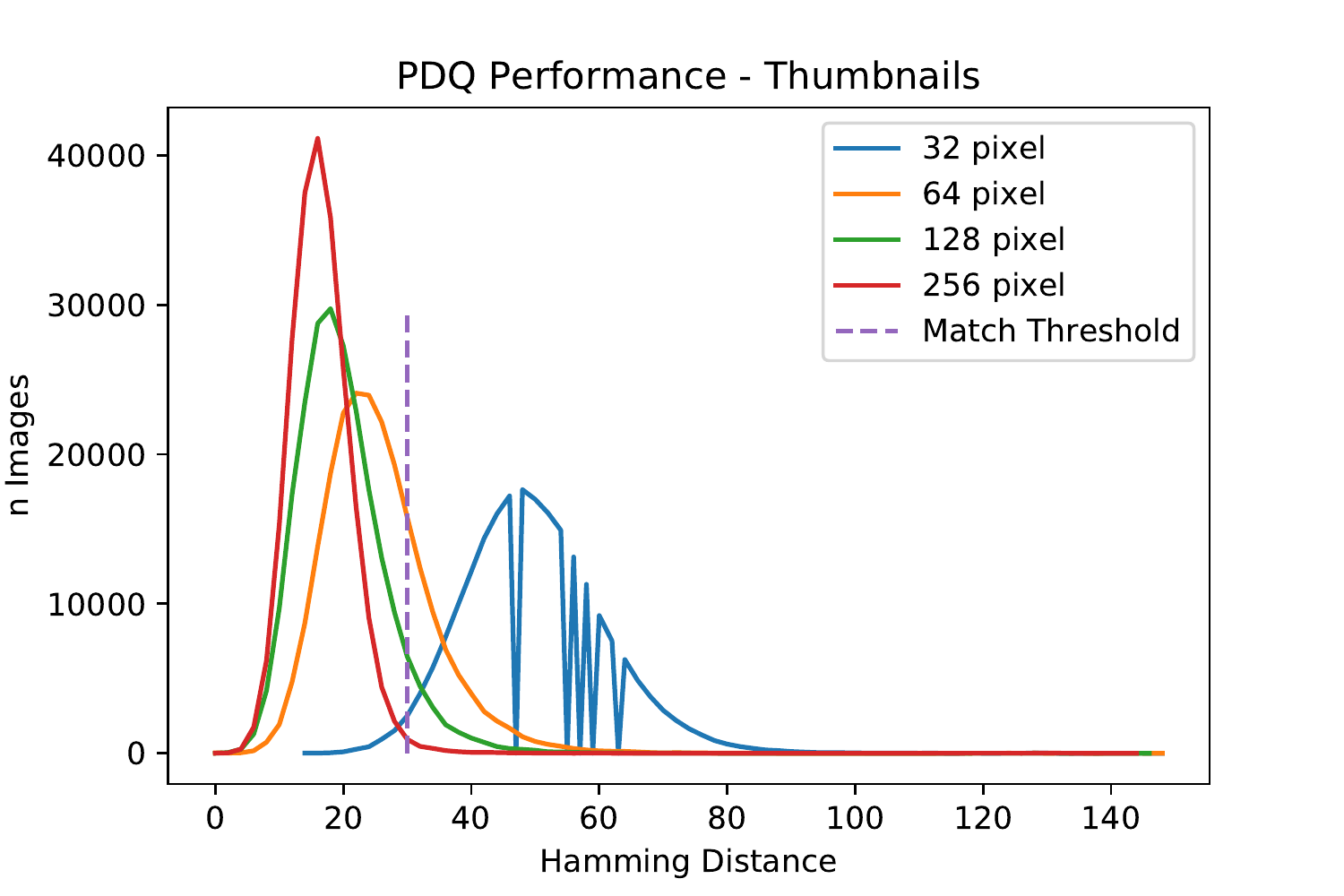}
\caption{Hamming distances for generated thumbnails}
\label{PDQThumbnailPlot}
\end{figure}

As with format changes, the algorithm performed strongly when responding to what really constitutes a loss in resolution. Figure \ref{PDQThumbnailPlot} displays that using the 30 hamming distance ceiling would almost guarantee detection of all 256 pixel (long side) images as duplicates of their originals. The loss of granularity is reflected in the results for the 128 and 64 pixel forms - both would detect most instances, but the latter definitely wouldn't do so for many. 

The 32 pixel thumbnail plot demonstrates the limitations of perceptual hashing when dealing with low quality imagery, with the series diving and spiking as it encounters odd numbered hamming distances. As discussed in Section \ref{PDQIntro}, quality images result in a PDQ hash containing an even number of 0 and 1 values, leading to a far higher likelihood of even hamming distances. As the sample in Figure \ref{sampleWatermark} shows, the quality of a 32px thumbnail is quite low, legible content in such a format is extremely limited.

\subsection{Cropping}
The algorithm struggled with cropping, and itself isn't designed to cope with such changes - instead, additional hashes can be generated to provide what amounts to hypothetical changes, including rotation. This option is available within the binary used, and was utilised for this test and for rotation.

\begin{figure}[h]
\includegraphics[width=\columnwidth]{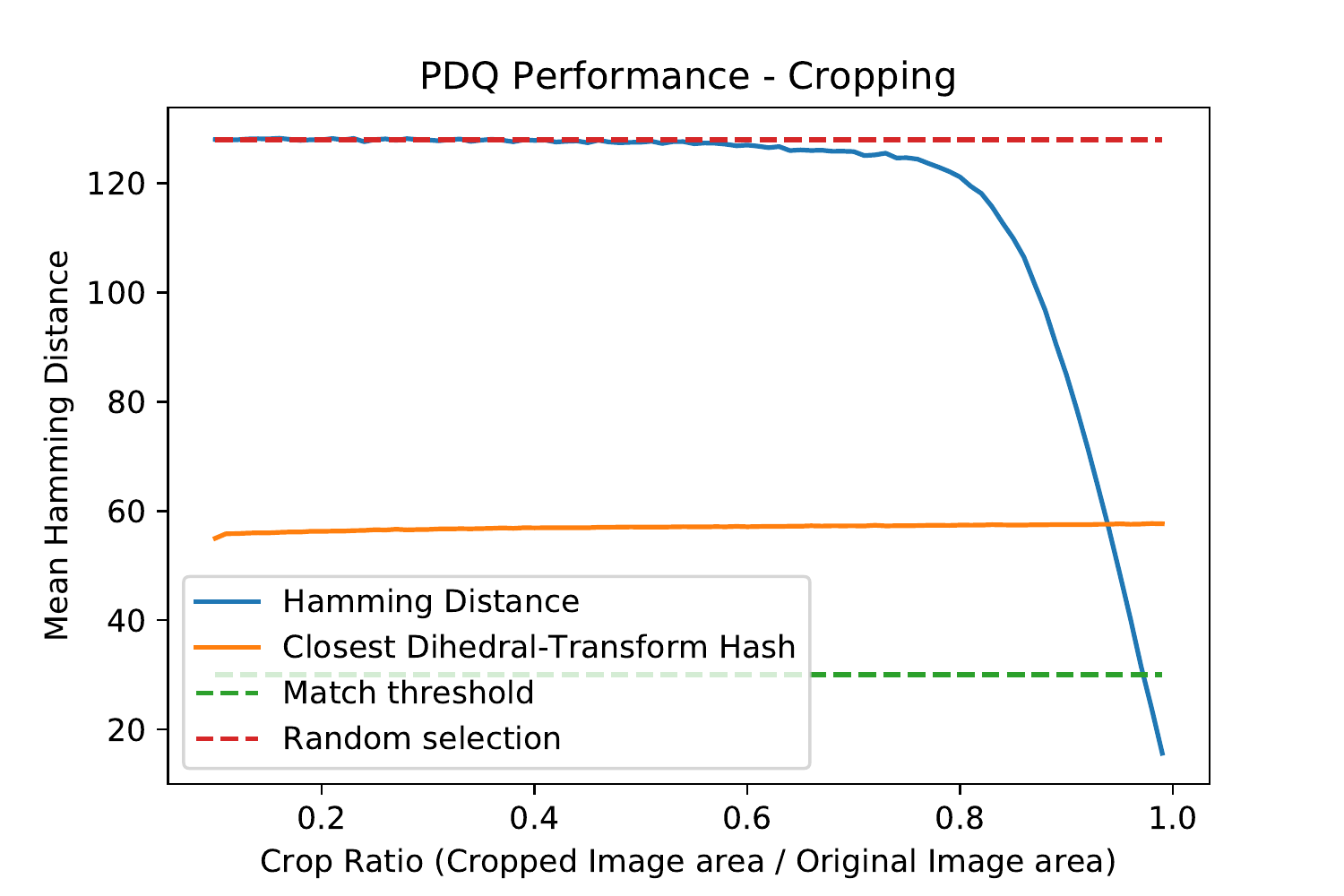}
\caption{Hamming distances for cropped images}
\label{PDQCropping}
\end{figure}

Figure \ref{PDQCropping} displays that a small ($<$5\%) removal of an image will result in hamming distances beyond our working threshold, with results nearing random distributions by around 20\% removal (i.e. a crop ratio of 80\%). When the Dihedral-Transform hashes are utilised, a near average of around 57 for the closest hash is returned across the cropping spectrum, though the use of these additional hashes would lead to (a) query time inefficiencies and (b) a far greater change of false positives. Of course, we would also need to set a threshold nearing 60 to utilise this approach, again increasing the likelihood of false positives.

\subsection{Rotation}
The algorithm's performance largely mirrored that observed for cropping. Beyond slight (less than 5\degree of original), performance falls away sharply with only calculation of additional hashes capable of reducing distances away from near-random distributions. As discussed, such an approach would be a good backstop if rotated/cropped images are anticipated, but otherwise would probably lead to undesirable performance impacts.

\begin{figure}[h]
\includegraphics[width=\columnwidth]{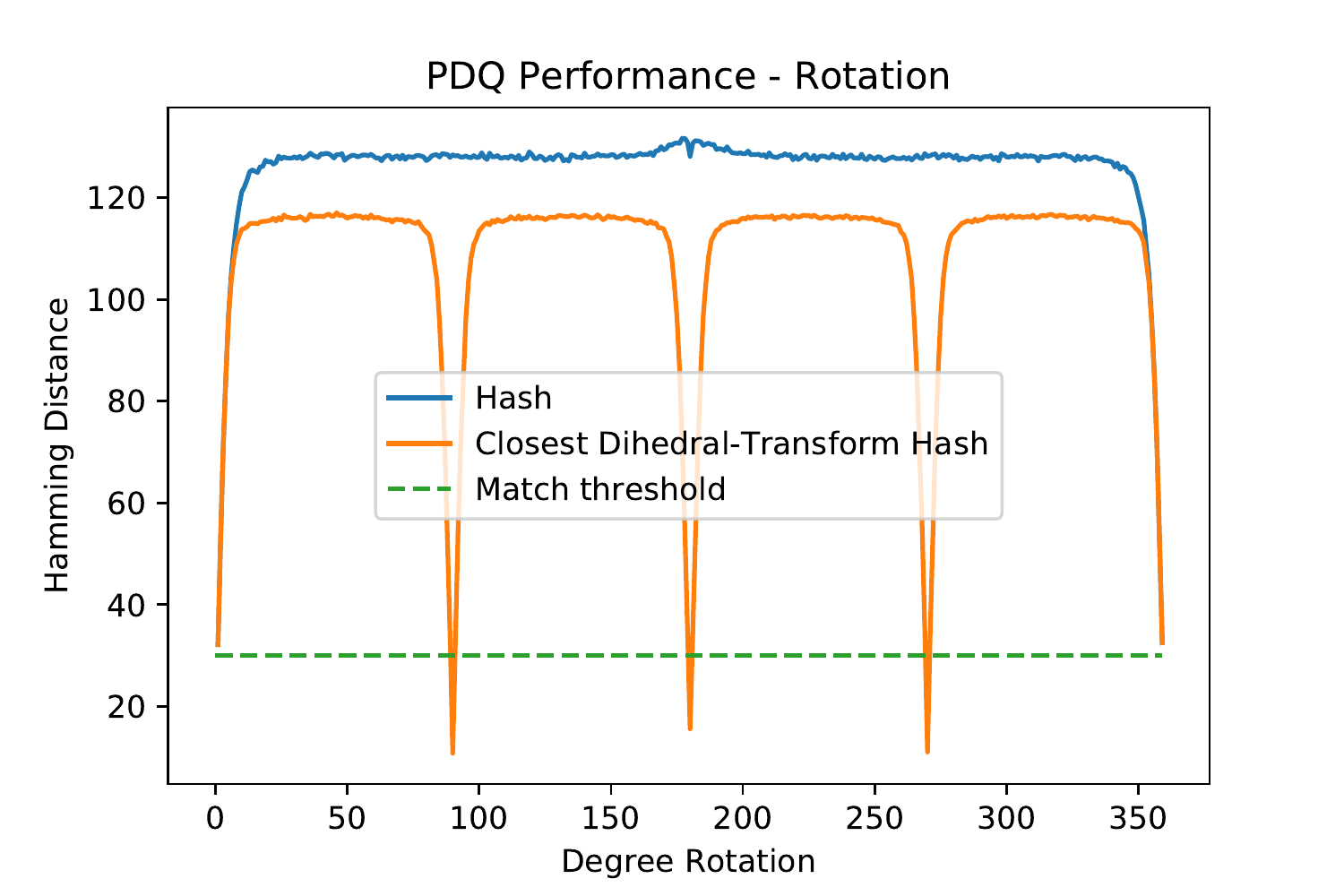}
\caption{Hamming distances for rotated images}
\label{PDQRotation}
\end{figure}

\subsection{Speed}
\begin{figure}[h]
\includegraphics[width=\columnwidth]{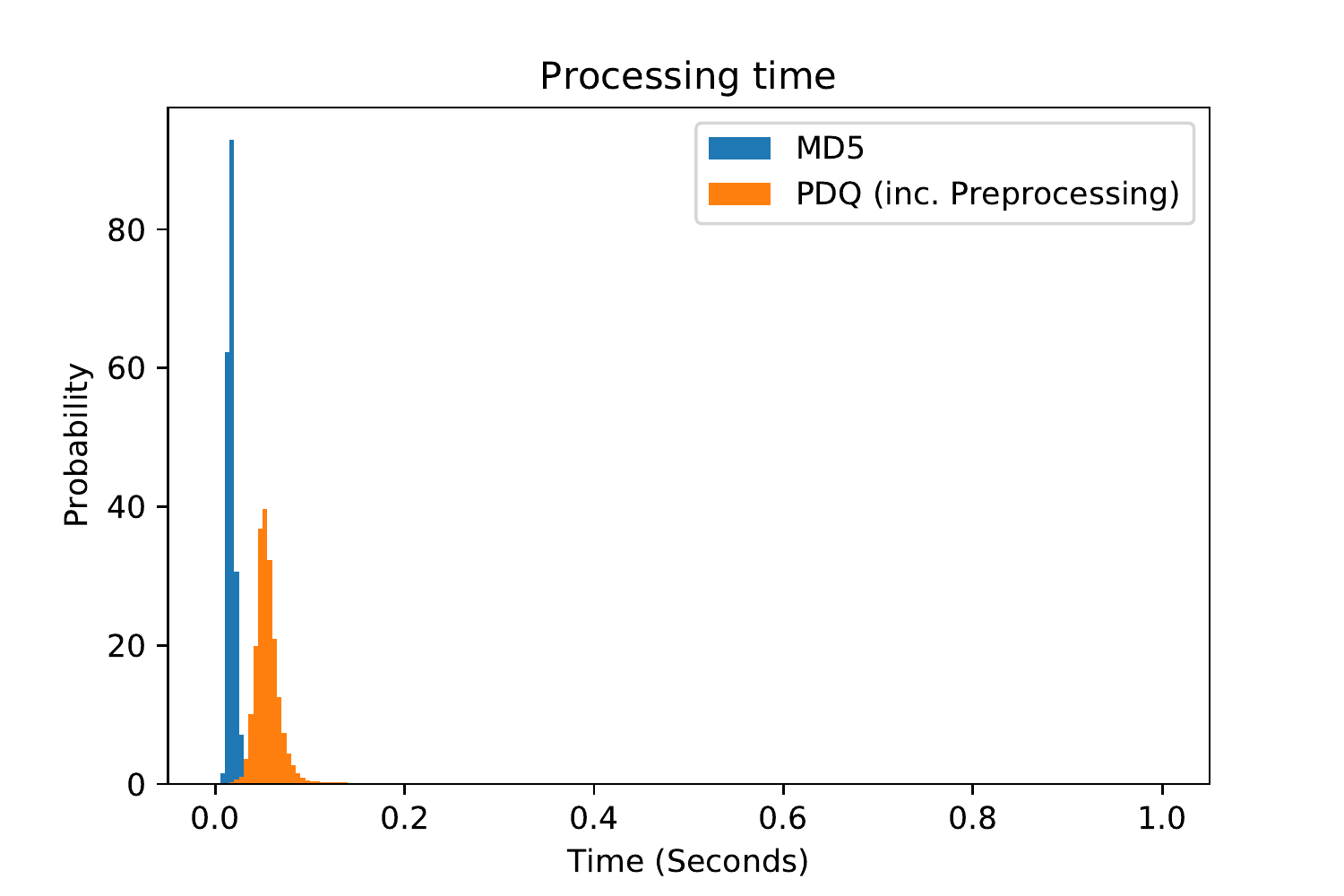}
\caption{Time taken for PDQ \& MD5 hash calculations}
\label{PDQvsMD5}
\end{figure}

PDQ hash generation involved a time overhead compared to that required for MD5. Figure \ref{PDQvsMD5} displays that whilst MD5 digests could be calculated for 90\% of files within 0.02 seconds, PDQ was largely completed around 0.08. This is a very reasonable amount considering PDQ is a perceptual hash (and therefore requires image rendering rather than simple binary reads and calculations). We did \textbf{not} pre-process imagery to accelerate the process, keeping all data 'as is' for processing. This performance can therefore be regarded as close to worst-case for real-world use, particularly considering the use of an external binary rather than calling integrated code.

\subsection{Entropy}
A fuzzy hash's values across a sufficiently large corpus of dissimilar materials should resemble random distribution. Such a `flat' distribution of values indicates a lack of bias (intentional or otherwise) within the algorithm, and in this instance, supports the use of a simple edit distance calculation for measuring perceived similarities. 

\begin{figure}[h]
\includegraphics[width=\columnwidth]{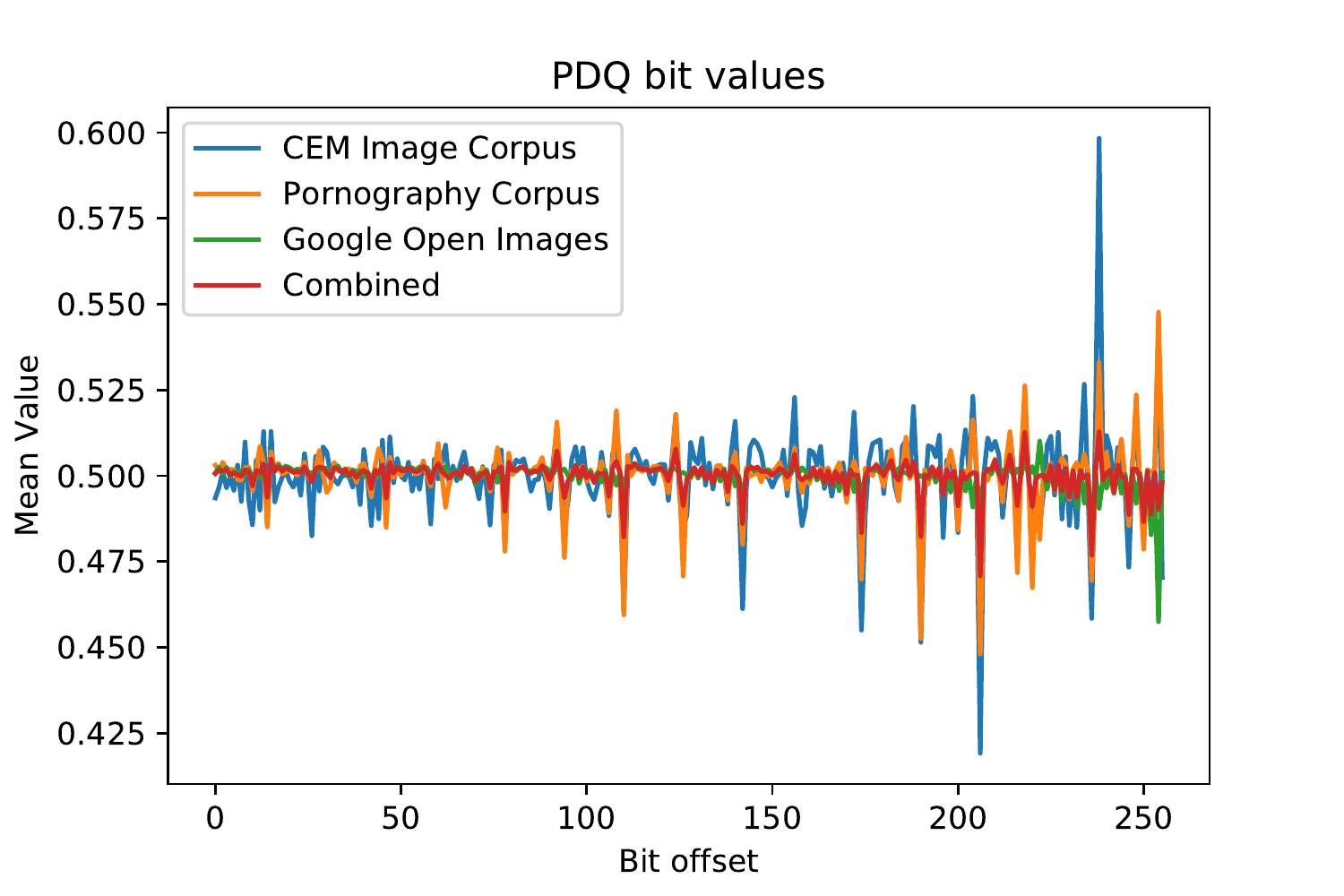}
\caption{Average (mean) bit values for PDQ hashes across multiple corpora (duplicate MD5 and PDQ values removed)}
\label{MeanPDQBits}
\end{figure}

Figure \ref{MeanPDQBits} displays the mean bit values for each PDQ hash generated over each of our test corpora, plus across \textit{all} test corpora. The plot contains some peaks, indicating some biases - particularly within the CEM corpus. A degree of bias is to be anticipated on a per corpus basis, given the propensity for certain features (e.g. head/shoulders against a flat background for passport photos) to appear in specific genres. This hypothesis is supported by the smoother plot for the combined corpora series. 

\subsection{Discussion}
We found PDQ to be a robust and capable algorithm for ``non-adversarially'' (in the words of the project authors) altered imagery. In particular, we found performance when dealing with resized or converted images to be near perfect, as demonstrated by Figure \ref{PDQFormatChanges}'s sparseness - the worst result being $\frac{1}{15}$ of an acceptable match threshold away from zero.

As advised in project documentation, image rotation and cropping result in greatly decreased performance, and given their impact on materials shown, should reasonably be regarded as adversarial - in other words, being undertaken in order to defeat detection, rather than any other purpose. 

The use of additional (Dihedral-Transform) hashes can help combat this, but our tests demonstrate how their effective use would require a substantial change in match thresholds. Not discussed yet is the impact this would have on query times, a topic we cover later in this paper.

\section{Testing Video (TMK + PDQF)}
As with PDQ, we utilised executables packaged with the project for our tests - in this case, \texttt{tmk-hash-video} for hash generation. Unlike PDQ, given the relative complexity of calculating similarities, we used the packaged binary (\texttt{tmk-two-level-score}) for hash comparison. 

Unlike PDQ, the TMK algorithm takes a two phased approach - if the first phase passes a pre-defined match threshold, then the second phase is attempted. If both results are \textbf{higher} (note contrast to PDQ) than the threshold, then a video is regarded as a match. For our tests, we followed the recommended threshold of 0.7 for both phases.

Note that \texttt{tmk-two-level-score} does not output scores if no match is detected for phase 1 - hence, setting the threshold to -1 in effect forces full calculations.

\subsection{Methodology}
We undertook a series of transformations akin to those utilised in testing PDQ, but adapted for use in video formats. The tests consisted of:
\begin{enumerate}
\item{Bitrate alteration - the visual bitrate (if applicable) or overall bitrate of the video is reduced to a randomly selected ratio beween 0-100\%, exclusive.}
\item{Cropping - as with still images, a randomly selected ratio in the range 0-100\% (exclusive) of the original vision is retained, centered on the existing visual centre and maintaining the existing aspect ratio.}
\item{Format change - the video is converted to a random selection of 'mpg', 'mp4', 'flv', 'mkv' and 'avi' formats.}
\item{Half scaling - effectively the video equivalent of thumbnailing, though in this case rather than following pre-defined sizes, the video's existing visual dimensions are halved.}
\item{Text - `AiLECS' (white, 40pt font) is scrolled across the screen for ten seconds, commencing one second into the video. This is repeated every 20 seconds thereafter.}
\item{Title - a five second title screen (consisting of the afp logo) is added to the video's beginning. No content is overwritten.}
\item{Trimming - five seconds of content are removed from the beginning of the video.}
\item{Watermark - The AFP logo was added to vision, following the same placement and sizing rules detailed within Section \ref{pdqmeth}.}
\end{enumerate}

\subsection{Results}
\begin{figure*}
\centering
\includegraphics[width=\textwidth]{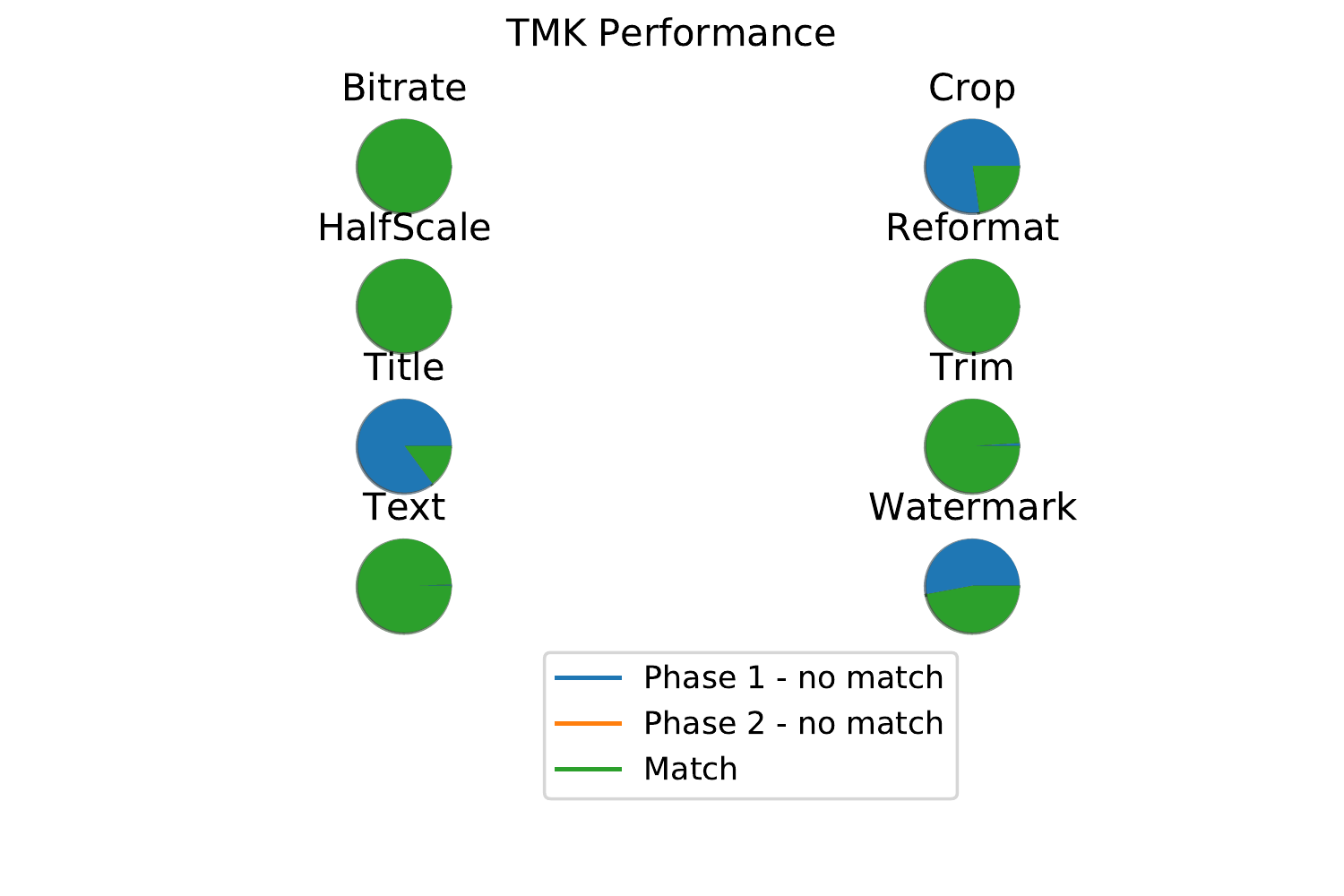}
\caption{Pass/fail rates for all video tests - Phase 2 assumes pass at Phase 1.}
\label{OverallTMK}
\end{figure*}

Figure \ref{OverallTMK} displays the overall performance of the TMK algorithm across our tests. As with still imagery (PDQ), the algorithm performs strongly when the perceived content changes are minimal - bitrate, scaling and format changes are all recognised as the `same' material. Watermark and cropping underperformed, akin to that seen with still imagery. Interestingly, adding a five second title screen also saw a dramatic loss of performance in recognising largely `like' materials.

\subsection{Bitrate alteration}
The algorithm performed extremely strongly with bitrate alterations, comfortably detecting each changed file as a match.

\subsection{Cropping}
\begin{figure*}
\centering
\includegraphics[width=\textwidth]{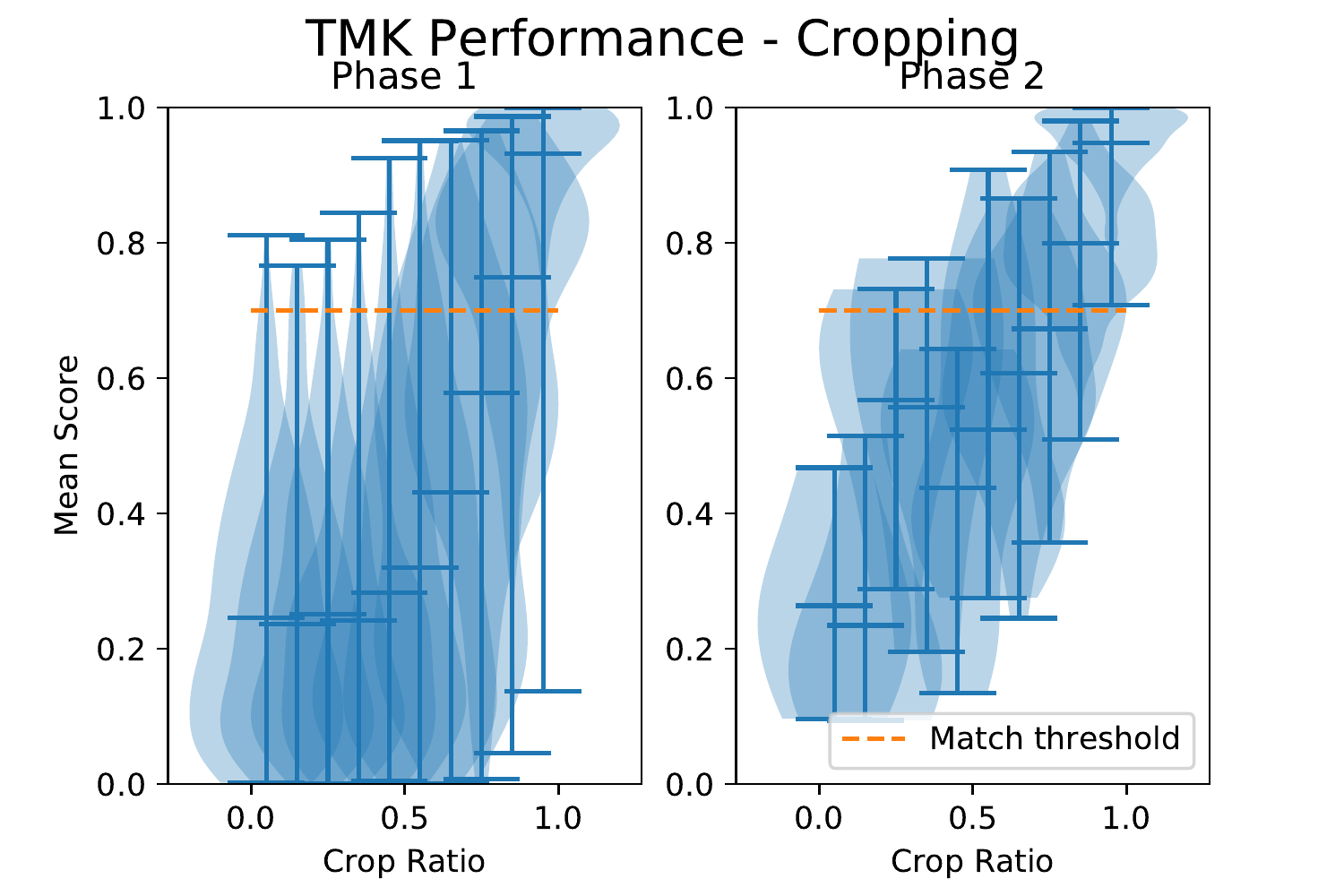}
\caption{TMK scores vs cropping ratios. Horizontal bars indicate max, mean and minimum values for each series.}
\label{CroppingTMK}
\end{figure*}

As with PDQ's performance on still imagery, the TMK algorithm struggled with cropping. Removing more than 20\% visuals made detection improbable during both phases, but as figure \ref{CroppingTMK} demonstrates, phase 1 performance drops rapidly in relation to crop ratio, until a floor of mean performance is reached around the 0.6 (i.e. removal of 40\%) level. Phase 2 follows this pattern, though with a more linear drop in confidence against cropping.

In an investigative context, we would recommend removal of more than 10\% be regarded as out of scope for reliable match detection.

\subsection{Format change}
Results in detecting content subjected for formatting changes closely followed those for bitrate alteration, with all files comfortably detected as duplicate content.

\subsection{Half Scaling}
Results for scaling closely followed those for bitrate and format alteration, though with marginally higher confidences within the phase 1 test. All, however, are comfortably above the match threshold of 0.7 and would only affect cases where extremely high thresholds are undertaken.

\subsection{Text} 
\begin{figure}
\centering
\includegraphics[width=\columnwidth]{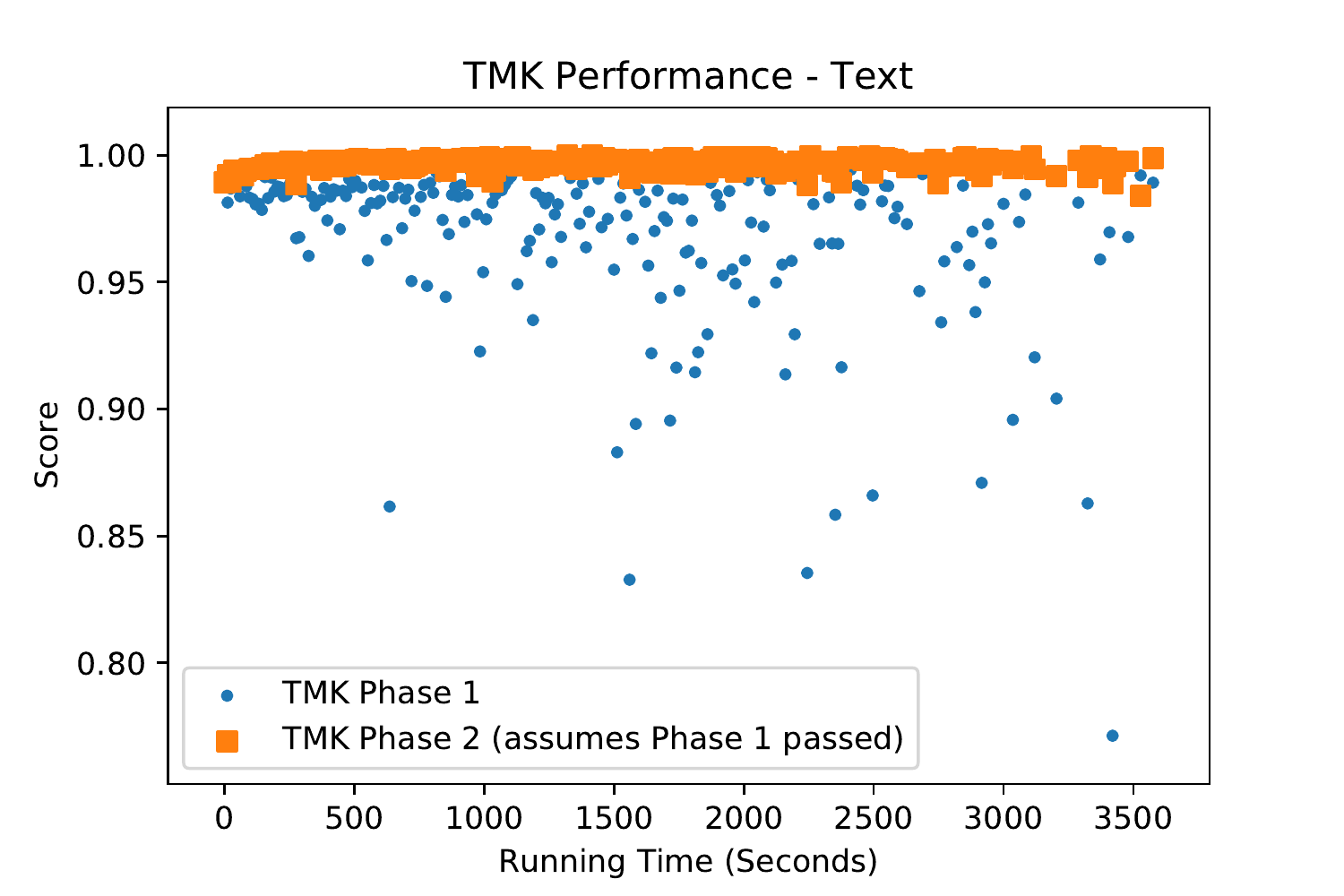}
\caption{Mean TMK scores for text test, taken against video running time. Note low correlation of scores to runtime, particularly within phase 1.}
\label{TMKTextRuntime}
\end{figure}

Less visually intrusive than the watermark, the algorithm was largely successful in detecting duplicate content. Whereas we assumed shorter runtimes could result in lower confidences (given the proportionately higher level of visual change), Figure \ref{TMKTextRuntime} shows this not to be the case. 

\subsection{Title}
\begin{figure}
\centering
\includegraphics[width=\columnwidth]{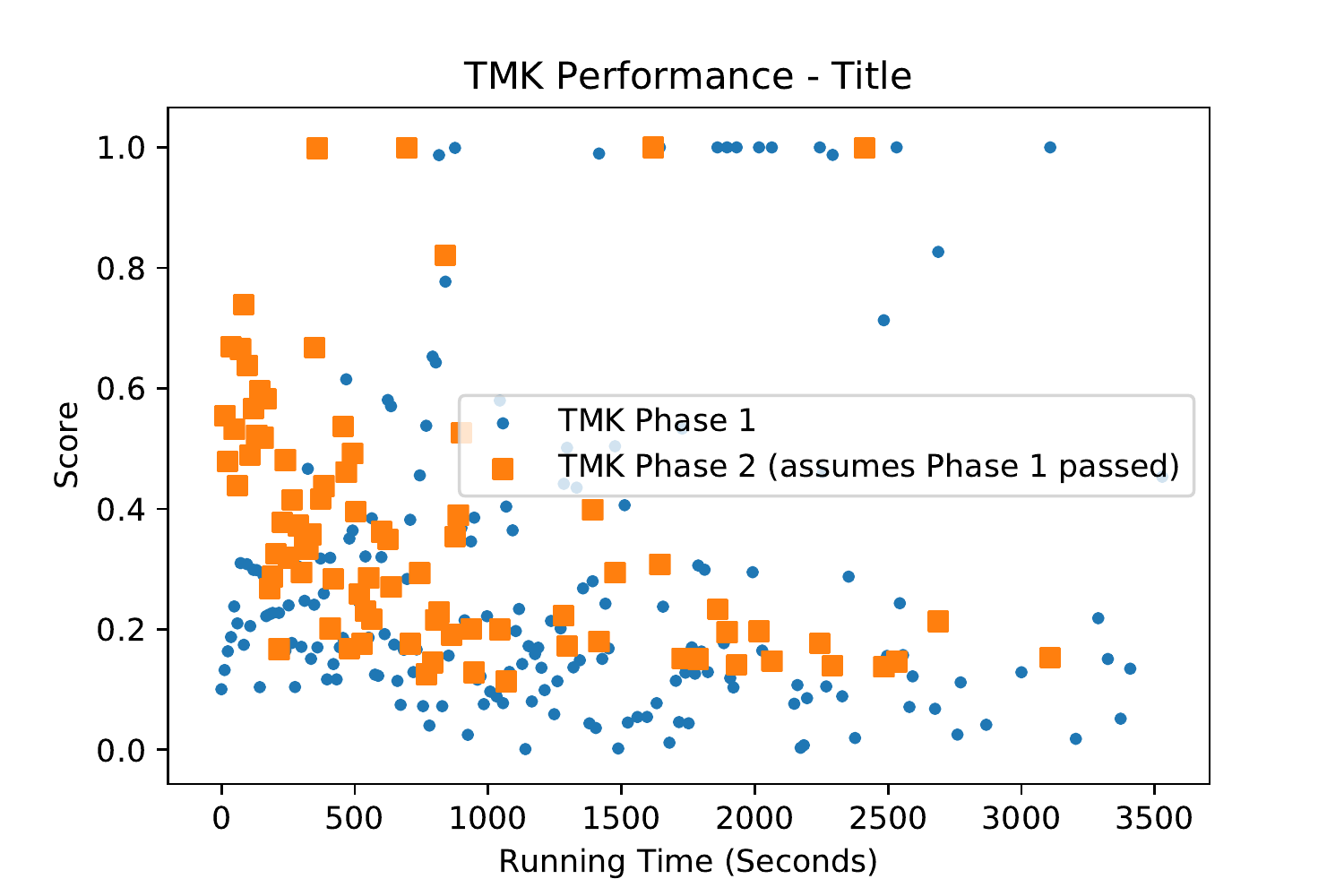}
\caption{Mean TMK scores for title test, taken against video running time.}
\label{TMKTitleRuntime}
\end{figure}

Overall, performance was disappointing in this test, with Figure \ref{TMKTitleRuntime} showing the majority of results falling below the 0.7 threshold for both phases. 

We originally hypothesised that the impact of new materials' insertion into a single location within a video would decrease as its relative proportion to running time decreased - in other words, the smaller the proportion of a video being new, the lower the change. It is difficult to confirm or disprove this theory. Prima facie, Figure \ref{TMKTitleRuntime} shows this not to be the case, as the clustering of Phase 2 scores (indicating a corresponding pass at Phase 1) occurs around $<= 1000$ seconds. This, however, is in line with the underlying distribution of runtimes within the test corpus. The absence of Phase 2 scores at longer running times could simply reflect a seemingly random distribution of successful Phase 1 tests, given the relative scarcity of sample data.
 
\subsection{Trimming}
\begin{figure}
\centering
\includegraphics[width=\columnwidth]{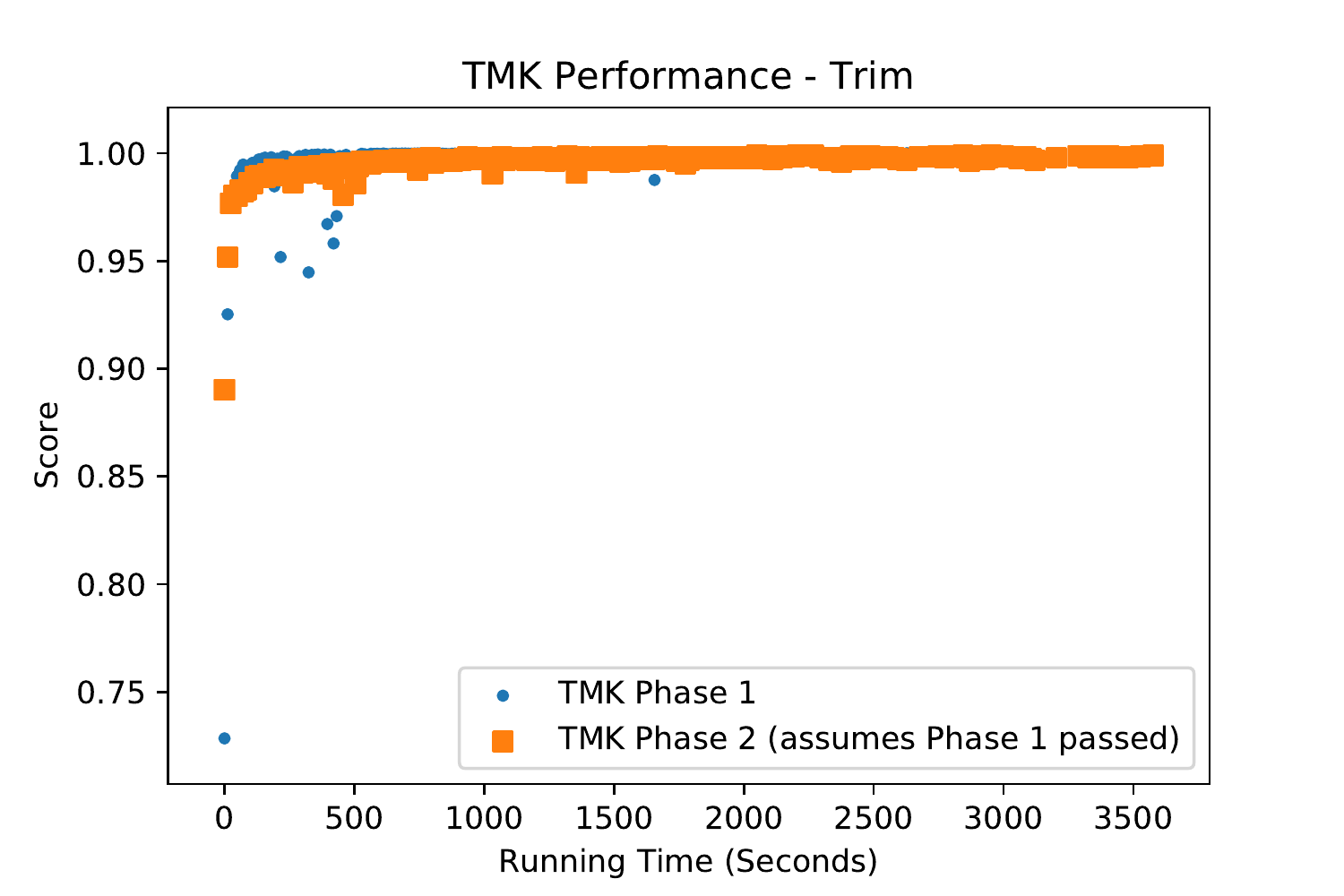}
\caption{Mean TMK scores for title test, taken against video running time.}
\label{TMKTrimRuntime}
\end{figure}

The TMK algorithm's performance in detecting `like' materials after the \textit{removal} of content stands in stark contrast to addition. Figure \ref{TMKTrimRuntime} displays the results - reversing the hypothesis given in the previous section, it would appear that performance does improve as the relative proportion of content altered (in this case, removed) decreases. This, however, occurs at a relatively short runtimes, with results approaching maximum at around 300 seconds. The only failure depicted on this plot is at an extremely short runtime, reflecting the substantial (if not entire) removal of content from that file.

\subsection{Watermark}
\begin{figure}
\centering
\includegraphics[width=\columnwidth]{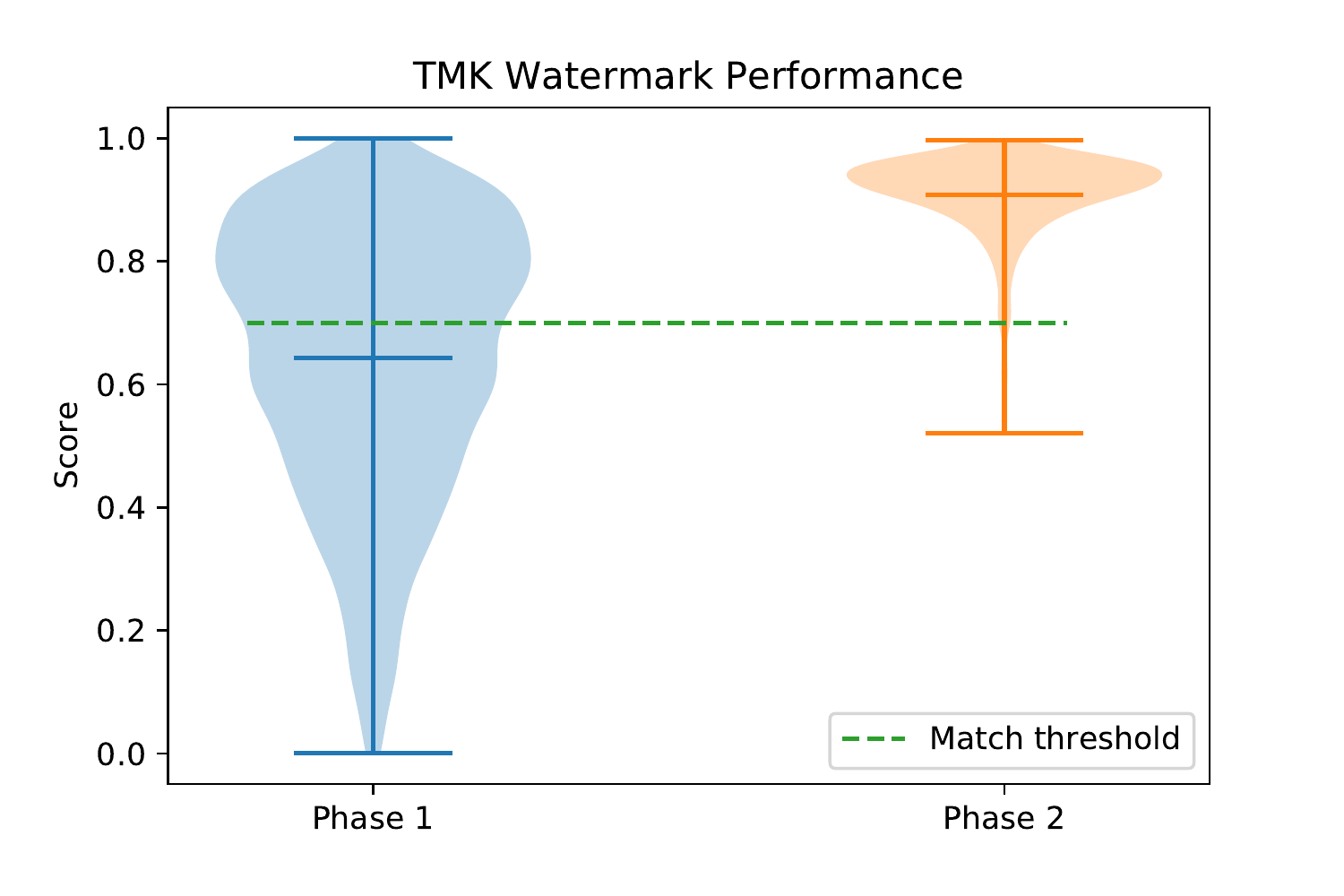}
\caption{TMK scores for watermark test. Phase 2 results only materials passing phase 1.}
\label{TMKWatermarkScore}
\end{figure}

Watermarking performance was similar for TMK as for PDQ. Figure \ref{TMKWatermarkScore} displays a mean score for videos falling below the recommended threshold of 0.7 within phase 1. On a more positive note, phase 2 performance was robust in cases where phase 1 was passed.

\subsection{Speed}
Whilst the wording is partly confusing, the project documentation lists TMK hashes as taking approximately 30x video runtime, depending on storage density. We experienced better speeds, with average hash calculation times being approximately $\frac{1}{40}$ (2.5\%) of runtime - probably due to our use of local (SATA3) based storage.

\begin{figure}
\centering
\includegraphics[width=\columnwidth]{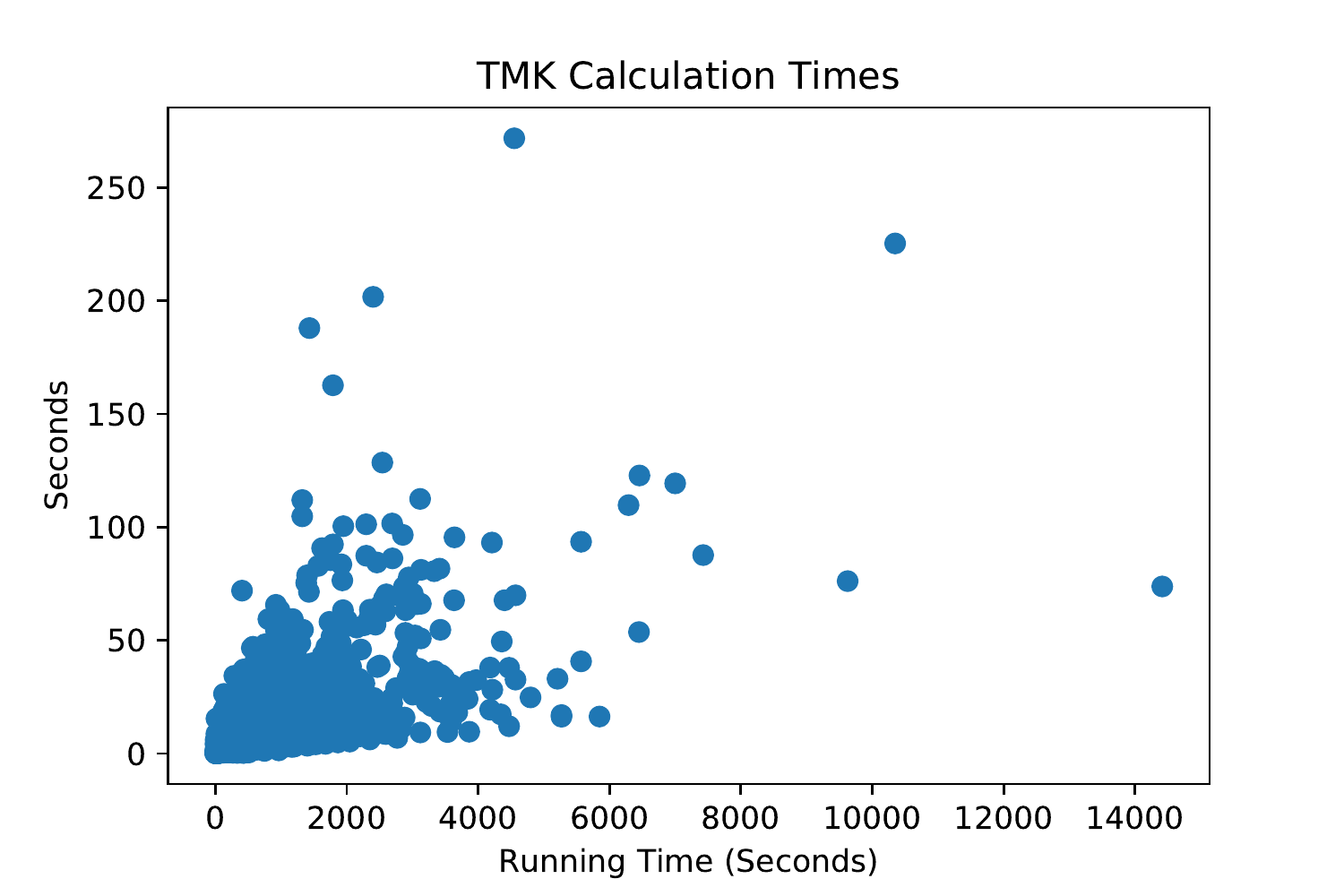}
\caption{TMK Calculation times vs video running times}
\label{TMKRuntime}
\end{figure}

Figure \ref{TMKRuntime} shows the correlation, though with variability most likely introduced by varying file sizes/compression levels and any required conversion to supported video formats.

\subsection{Entropy}
The storage and comparison of PDQ hashes are relatively simple, particularly given their compact size and simple format - 256 bit hexadecimal strings being readily transportable within a text file, or stored as fields within databases. In comparison, the TMK algorithm (as released) stores hashes as 256 kilobyte file - approximately 8,000 times larger. The authors make a note of the first 1 KB being sufficient for differentiating most videos, but we are unaware of any firmer statistics regarding such performance.

\subsection{Discussion}
We observed the TMK + PDQF algorithm to work as stated within project documentation. It is a robust and well performing algorithm in the uses for which it was designed - format, resolution and compression changes result in near perfect performance, closely followed by `light-touch' alterations such as text. Heavier, ``adversarial'' changes (as described by the project authors) such as the addition of a title screen, watermarking and cropping beyond marginal levels result in rapid performance drops.

\section{Implementation Discussion}
Beyond integration and licencing concerns, arguably the greatest roadblock to the widespread law enforcement adoption of perceptual hashing is query time performance. Unlike cryptographic digests such as MD5 and SHA-1 where the smallest non-zero hamming distance and maximum value are equivalent, fuzzy hashes such as PDQ require comparison of a large proportion (if not entirety) of each value to establish similarity. This is substantially more computationally intensive than comparing cryptographic digests - in the case of MD5, 99.6\% $(1 - \frac{1}{2^{8}})$ of digests can be disregarded after the eighth bit\footnote{Second character if displayed as a hexadecimal string}. Assuming a threshold of 30, a search for matching images using PDQ requires comparison of at least 30 bits (7.5 hex characters), almost four times more than the overwhelming majority of MD5 digests.

The PDQ documentation includes advice and mathematical proofs on the feasibility of advice on indexing methods such as multi-index hashing (MIH) \citep{6248043} for minimising query times. We have incorporated MIH as part of our reference implementation (refer Section \ref{Implementation}), observing a drop in query times of about 33\% when compared with linear search across our test corpus, a result becoming more pronounced as the corpus size increases.

Unlike the PDQ authors, we dropped the use of caching within the index, finding the overall speed improvement within our tests to be marginal. This could be due to our use of python rather than a pre-compiled executable, and in any case, cache performance is entirely dependent upon individual usage patterns. Our implementation of MIH is available for download at \url{https://github.com/AiLECS/pyMIH} or via \texttt{pypi/pip} as \texttt{pyMIH}.

As described earlier in this paper, TMK is also capable of accelerated performance through indexing, with the relatively lightweight outputs of Phase 1 ample for rapidly reducing the number of candidates at lookup. Facebook has also open-sourced \texttt{faiss}\citep{Faiss}, an efficient search/indexing project with great potential for performance improvements. At time of writing, integration between TMK and faiss is faulty. A workaround has been provided, with a fix to transparently avoid the issue reportedly in the works. 

\subsection{Reference Implementation}
\label{Implementation}
Indexing can improve performance in fuzzy hash lookups, though it will never replicate the speed of cryptographic hashes - hence their continuing use, despite well known limitations. Computational overheads are acceptable in smaller datasets, but grow exponentially as large-scale databases develop and grow. On-demand (i.e. `cloud') computing is one step towards overcoming delays, either by providing more resources, or additional resources closer to the edge - i.e. where the demand is. Given existing dependencies, the PDQ binaries are currently only deployable on *nix and MacOS systems, precluding its use on a vast majority of corporate systems. We have packaged the PDK implementation as a Docker image\footnote{Available for download at \url{https://github.com/JDalins/PDQContainer}}, exposing the algorithm via a RESTful API service for accessibility. Whilst containerisation doesn't accelerate computation, it massively improves portability, allowing the service to spin up and down across heterogeneous host operating systems and infrastructure. The image itself is 1.2GB, making it easily transported over standard internet connections. 

A TMK image is under construction, but due to the additional search overheads, will not be released until a Faiss-based search back-end is completed. We anticipate release of this image by mid-late October 2019. Further development of both images will be dependent on performance and user demand. 

%\todo{This is out of scope - happy to remove}
%An alternate approach to dealing with demand is to change usage patterns. Most forensic tools within law enforcement carry out lookups as part of a batch ingestion process - for digital forensics, at time of seizure. This is acceptable in situations of uniform demand, but this is a usage pattern rarely seen within policing. Instead, moving more expensive services to a `just in time' model (i.e. a forensic tool calls individual services in anticipation of user review) could significantly smooth demand spikes, with less urgent, routine tasks rescheduled to off-peak times. 

\section{Discussion}
A common lament within digital forensics (and law enforcement more broadly) is the gap between R\&D and implemented, accessible tools. Microsoft's free (in dollar terms) release of PhotoDNA was an unprecedented step at the time, but its ongoing status as a proprietary algorithm limits its further distribution, evolution and application. 

The release of free software, both in terms of monetary cost and restrictions on application, significantly reduces the non-technical overheads restricting implementation of novel technologies into existing forensic tools and workflows. In our case, a simple git clone followed by two `make` commands was sufficient for establishing a basic tool - the entire process taking less than twenty minutes. Development of the python script for automating testing took a further two days, making this rudimentary deployment a weekend project. 

In its current form, external dependencies limit the project's deployment in environments other than *nix and MacOS - the AFP (as with many other LEAs) is predominantly a Windows environment, with in-house software development typically gravitating to .NET. The ability to call hashing algorithms directly from existing software would represent a massive leap in the deployment potential for tools such as these. .NET wrappers or ports \footnote{Most likely in the form of a dll}, released and supported as part of the project, would certainly simplify deployment of these algorithms within such environments.

Our use of the PDQ and TMK algorithms is basic, with scripts calling command line applications rather than integrating tightly with our tools. Substantial performance increases could be achieved through tighter coupling of libraries with host applications - the ability to directly pass image/video data in memory (rather than via writes to file systems) alone would accelerate performance dramatically. 

The AFP is moving to a loosely coupled microservices based architecture - primarily to ensure agility in delivering the tools demanded by different crime threat domains, but also ensuring their availability and adaptability across the organisation and beyond. The deployment of services such as image/video similarity measurement via a container exposing a RESTful API accelerates both implementation and use: Specialist client applications such as digital forensic tools can utilise basic HTTP libraries for the vast bulk of integration work, whereas containers can be reliably orchestrated and rapidly deployed to areas of demand, reducing network overheads (a critical concern given Australia's geography). Our reference implementation of such a service demonstrates the concept's viability, though it must be noted that the underlying database deployment for these microservices is going to be a matter for individual organisations.

In this review we have identified the potential value the PDQ and TMK algorithms bring to law enforcement, particularly in reducing repeat practitioner exposure to known (but even imperceptibly altered) materials. The open sourcing of such algorithms presents a major (if not unprecedented) opportunity for LEAs and their partners to not only use and share leading-edge tools, but also guide and contribute to their development and evolution. The initiative now rests with practitioners to take the intitiative provided by releases such as this, and where suitable, continue the development of such tools into deployable and sustainable products. 

\section{The AiLECS Lab}
The work depicted within this paper was undertaken as part of the Artificial Intelligence for Law Enforcement \& Community Safety (AiLECS) Lab, a joint Australian Federal Police/Monash University initiative formed as part of Monash University's "IT for Social Good" program. 

The lab's mission is to research and develop ethical uses of machine learning within law enforcement, with a particular focus the use of advanced data analytics in the identification and classification of psychologically harmful materials. For more information, please refer to \url{https://www.monash.edu/it/ailecs}.

\bibliography{tmkReport}
\bibliographystyle{ieeetr}
\end{document}